\documentclass[runningheads]{llncs}

\usepackage{accv}

\usepackage{accvabbrv}

\usepackage{graphicx}
\usepackage{booktabs}
\usepackage{wrapfig}
\usepackage{microtype}
\usepackage{marvosym}

\usepackage[accsupp]{axessibility}  % Improves PDF readability for those with disabilities.

\usepackage{colortbl}
\usepackage[table]{xcolor}
\usepackage{hyperref}

\usepackage{orcidlink}

\begin{document}

\title{Risk-Aware Selective Multimodal Driver Monitoring with Driver-State World Modeling}

\titlerunning{Risk-Aware Selective Multimodal Driver Monitoring}

\author{Daosheng Qiu\inst{1}\orcidlink{0009-0002-2855-5254} \and
Haozhuang Chi\inst{2}\thanks{Project Lead.}\orcidlink{0009-0000-7951-1046} \and
Hao Su\inst{3}\orcidlink{0009-0004-7679-7842} \and
Shu Long\inst{1}\orcidlink{0009-0009-1423-8275} \and
Xinyue Miao\inst{1}\orcidlink{0009-0001-6471-5350} \and
Yongle Dong\inst{1}\orcidlink{0009-0005-5298-232X} \and
Wei Zhang\inst{1}\textsuperscript{\Letter}\thanks{Corresponding author.}\orcidlink{0009-0009-8178-230X
}
}

\authorrunning{D.~Qiu et al.}

\institute{Hubei University, Wuhan, China \and
Nanyang Technological University, Singapore\and
Osaka University, Osaka, Japan\\
\email{zhang\_wei@hubu.edu.cn}}

\maketitle

\begin{abstract}
Continuous driver monitoring in automated vehicles requires low-latency inference while avoiding unsafe decisions under uncertain driver states. Large vision-language models provide broad multimodal priors, but their latency and limited reliability in this setting make them unsuitable as always-on in-cabin monitors. We propose a cost-aware selective inference framework for deployable multimodal driver monitoring. The core system is a lightweight RGB-physiological student that combines in-cabin visual observations with window-level HR/EDA signals, and a learned gate that decides when to accept the fast prediction or abstain for safety intervention. Additional controls show that the learned scores contain sample-level information beyond scenario priors, while exact physiological synchronization remains a limitation. To incorporate predictive evidence, we further study a compact driver-state world modeling module that rolls out latent driver-state features and estimates future fast-model errors and counterfactual system-level action costs. On scenario-induced driver-demand recognition, the RGB-physiological student improves over RGB-only and physiology-only baselines, reaching 0.7440 Macro-F1 and 0.9099 balanced accuracy with 11.39M parameters and 3.08ms inference latency. Cost-aware selective inference reduces unsafe false negatives from 17.37\% under always-fast inference to approximately 5\% across seeds, while maintaining deployment-level latency. While driver-state world modeling offers valuable predictive signals, worst-group evaluations highlight persistent operating-point calibration drift. Ultimately, reliable edge driver monitoring requires advancing not only perception backbones, but also risk-aware selective control and group-robust calibration.


\end{abstract}

\section{Introduction}
\label{sec:introduction}

Human-centered driver monitoring is a prerequisite for safe automated driving, especially during shared-control and control-transition scenarios. A monitoring system must operate continuously, respond with low latency, and avoid unsafe decisions when the driver's state is ambiguous. This requirement is particularly important for demand-related driver states, which are often only partially observable from in-cabin visual cues. A driver may experience elevated demand without exhibiting a distinctive facial expression, posture, or hand movement, while window-level physiological signals such as heart rate and electrodermal activity can provide complementary evidence related to the driver's state.

These requirements create a tension between multimodal understanding and deployability. Recent vision-language models (VLMs) provide broad visual-semantic priors and offer a flexible interface for multimodal reasoning. However, our empirical study shows that large VLMs are not well suited as always-on in-cabin monitoring models. Their latency is orders of magnitude higher than that of compact task-specific networks, and their reliability as direct predictors or distillation teachers is limited under the driver-demand recognition protocol considered in this work. This motivates a different design principle: the continuous monitoring path should be handled by a lightweight multimodal student, while heavier predictive or verification modules should be invoked only selectively when they provide useful risk-related evidence.

We propose a cost-aware selective inference framework for deployable multimodal driver monitoring. The core perception model is a lightweight RGB-physiological student that fuses short in-cabin visual observations with heart-rate and electrodermal activity signals. Instead of forcing the model to produce a decision for every input, we introduce a learned selective gate that evaluates prediction confidence, uncertainty, modality disagreement, physiological signal quality, and deployment cost. The gate decides whether to accept the fast prediction or abstain for safety intervention. This formulation separates perception accuracy from decision risk: in safety-critical monitoring, reducing unsafe false negatives can be more important than maximizing coverage on every sample.

To incorporate predictive evidence, we further study a compact driver-state world modeling module. Unlike scene-level world models for autonomous driving that generate or predict future external environments, our formulation operates in a latent driver-state space. It performs action-conditioned closed-loop rollouts to estimate future fast-model errors and counterfactual system costs. The module is designed to support selective monitoring rather than to replace the fast student. Our experiments show that driver-state world modeling provides useful future-risk and action-cost signals, but worst-group evaluations also reveal a persistent limitation: operating points selected on validation data may drift under heavy-tailed group shifts. This observation highlights group-robust calibration as a central challenge for predictive safety monitoring.

We evaluate the proposed framework on scenario-induced driver-demand recognition with RGB and physiological measurements. The RGB-physiological student improves over RGB-only and physiology-only baselines, reaching 0.7440 Macro-F1 and 0.9099 balanced accuracy with 11.39M parameters and 3.08 ms inference latency. Cost-aware selective inference reduces unsafe false negatives from 17.37\% under always-fast inference to approximately 5\% across seeds while preserving deployment-level latency. We further analyze VLM teachers, knowledge distillation, physiological reliability, driver-state world modeling, and worst-group robustness. The results suggest that reliable edge driver monitoring requires not only stronger perception models, but also selective risk control and calibration under group shift.

Our contributions are summarized as follows:
\begin{itemize}
    \item We introduce a deployable RGB-physiological driver monitoring framework and show, through scenario-prior probes, error-recovery analysis, and physiological intervention controls, that the learned scores contain sample-level evidence beyond RGB-only monitoring and scenario lookup.
    \item We formulate multimodal driver monitoring as a cost-aware selective inference problem, enabling the system to trade off coverage, latency, abstention, and unsafe false negatives.
    \item We study a compact action-conditioned driver-state world modeling module that performs latent rollouts to estimate future fast-model errors and counterfactual system costs for selective monitoring.
    \item We provide a systematic empirical analysis of VLM teachers, knowledge distillation, physiological reliability, deployment latency, and worst-group calibration, showing both the promise and limitations of predictive selective monitoring in safety-critical in-cabin sensing.
\end{itemize}

\section{Related Work}
\label{sec:related_work}

\subsection{Multimodal Driver Monitoring}
\label{sec:rw_multimodal_driver_monitoring}

Driver monitoring is central to safe assisted and automated driving, particularly in shared-control and takeover scenarios. Early systems commonly relied on visual indicators such as eye closure, gaze direction, head pose, facial or body motion, and other behavioral cues, while recent benchmarks have pushed the field toward deep in-cabin activity understanding. Drive\&Act \cite{Martin2019DriveAct} provides a large-scale multi-view dataset with RGB, infrared, depth, and pose modalities, together with hierarchical annotations for driver behavior recognition. DMD \cite{Ortega2020DMD} further broadens visual driver monitoring with large-scale RGB, depth, and infrared recordings for attention, alertness, and distraction analysis.

Vision-only monitoring remains limited when the target state is not directly observable. Demand-related states, stress, workload, and reduced readiness may occur without a distinctive visual manifestation. Physiological measurements provide complementary evidence in such cases. Prior work has shown that autonomic signals, including heart activity and electrodermal activity, are informative for stress, workload, and drowsiness-related driver-state estimation \cite{Healey2005Stress,Sahayadhas2012DrowsinessReview,Meteier2021Workload,Sriranga2023PhysioReview,Huang2024DriverStatePhysio}. The manD 1.0 dataset \cite{DargahiNobari2024manD} enables synchronized multimodal driver modeling with visual, physiological, gaze, motion, and vehicle-related measurements in assisted driving automation. Our work follows this multimodal direction, but focuses on a deployable RGB-physiological student and explicitly studies reliability, shortcut behavior, and selective decision-making for scenario-induced driver-demand recognition.

\subsection{Vision-Language Models and Deployable Students}
\label{sec:rw_vlm_deployment}

Vision-language models have established powerful general-purpose interfaces for visual-semantic understanding. CLIP \cite{Radford2021CLIP} demonstrated scalable language-supervised visual pretraining, while Flamingo \cite{Alayrac2022Flamingo}, BLIP-2 \cite{Li2023BLIP2}, InstructBLIP \cite{Dai2023InstructBLIP}, LLaVA \cite{Liu2023LLaVA}, and Qwen2-VL \cite{Wang2024Qwen2VL} advanced multimodal reasoning through large language models, instruction tuning, and unified image-video interfaces. Recent VLM-based driver-monitoring work \cite{VLMDM} has explored visual instruction models for multitask driver understanding.

Despite these advantages, large VLMs are not naturally suited for continuous driver monitoring. Their autoregressive decoding and large parameter counts impose substantial latency, and their reliability as task-specific predictors or distillation teachers is not guaranteed in safety-critical in-cabin settings. Knowledge distillation offers one route to transfer information from large models to compact networks \cite{Hinton2015Distilling}, and recent work such as VL2Lite \cite{Jang2025VL2Lite} studies task-specific distillation from VLMs to lightweight classifiers. In contrast, our empirical findings show that a slow VLM teacher is not consistently stronger than a supervised RGB-physiological student under our driver-demand protocol. This motivates a student-first design: the always-on path is a compact multimodal model, while the VLM branch is treated as a probe or optional verification component rather than as the continuous monitoring engine.

\subsection{Selective Prediction and Risk-Aware Inference}
\label{sec:rw_selective_prediction}

Selective prediction allows a model to abstain when its prediction is unreliable. The classical reject-option formulation studies the trade-off between recognition error and rejection \cite{Chow1970RejectOption}. Modern deep selective classification extends this idea to neural networks by optimizing coverage and risk jointly \cite{Geifman2017Selective}. SelectiveNet \cite{Geifman2019SelectiveNet} integrates a selection function into the model, while Deep Gambler \cite{Liu2019DeepGamblers} formulates abstention through a portfolio-inspired objective. Related learning-to-defer methods study when a model should hand off decisions to an external expert \cite{Madras2018Defer,Mozannar2020Defer}. More recent risk-control work extends this line by calibrating model behavior under user-specified loss functions \cite{Angelopoulos2024CRC}.

Safety-critical driver monitoring requires a more asymmetric view of risk. In this setting, a false negative on a high-demand or high-risk driver state can be more costly than abstaining and triggering a safety intervention. Moreover, the decision must account not only for classification confidence, but also for latency, modality reliability, and deployment cost. Real-world deployment further raises group-shift concerns: models can perform well on average while failing on atypical groups or shifted test distributions \cite{Sagawa2020GroupDRO,Koh2021WILDS}. Our framework therefore formulates multimodal driver monitoring as cost-aware selective inference. The learned gate evaluates confidence, entropy, margin, modality disagreement, physiological signal quality, and deployment cost to decide whether to accept the fast student prediction or abstain. We extend this view to deployment-aware monitoring with asymmetric unsafe-FN costs.

\subsection{Predictive State Modeling and World Models}
\label{sec:rw_world_models}

World models learn predictive representations of how states evolve over time and how actions influence future outcomes. Early work introduced world models as compact dynamics models for reinforcement learning agents \cite{Ha2018WorldModels}. PlaNet \cite{Hafner2019PlaNet} and Dreamer-style agents \cite{Hafner2023DreamerV3} learn latent dynamics for planning or policy learning through predicted future trajectories, while TD-MPC2 \cite{Hansen2024TDMPCTwo} performs model-predictive control in a learned latent space. In visual representation learning, V-JEPA \cite{Bardes2024VJEPA} shows that predicting future or missing representations in latent space can yield strong video representations without pixel reconstruction. In autonomous driving, generative world models such as DriveDreamer \cite{Wang2024DriveDreamer} and GAIA-1 \cite{Hu2023GAIA1} focus on modeling future external driving scenes. Recent efforts such as Driver-WM \cite{chi2026driverwmdrivercentrictrafficconditionedlatent} have begun to shift this paradigm inward, proposing traffic-conditioned latent world models specifically designed for rolling out in-cabin driver dynamics.

Our setting is different. We do not generate future road scenes or model the external traffic world. Instead, we study compact driver-state world modeling for in-cabin monitoring. The model operates on latent RGB-physiological features and estimates future fast-model errors and counterfactual system costs under candidate actions. Its purpose is to provide predictive evidence for cost-aware selective inference rather than to replace the fast student. This distinction is important: driver-state world modeling is used here as a risk-forecasting and decision-support component for selective monitoring, and our experiments further show that calibration drift under worst-group shifts remains a central challenge.

\section{Methodology}
\label{sec:method}

% ==================== 插入主架构图 ====================
\begin{figure}[t]
    \centering
    % 使用 \linewidth 确保图片撑满当前页面的文本宽度
    \includegraphics[width=\linewidth]{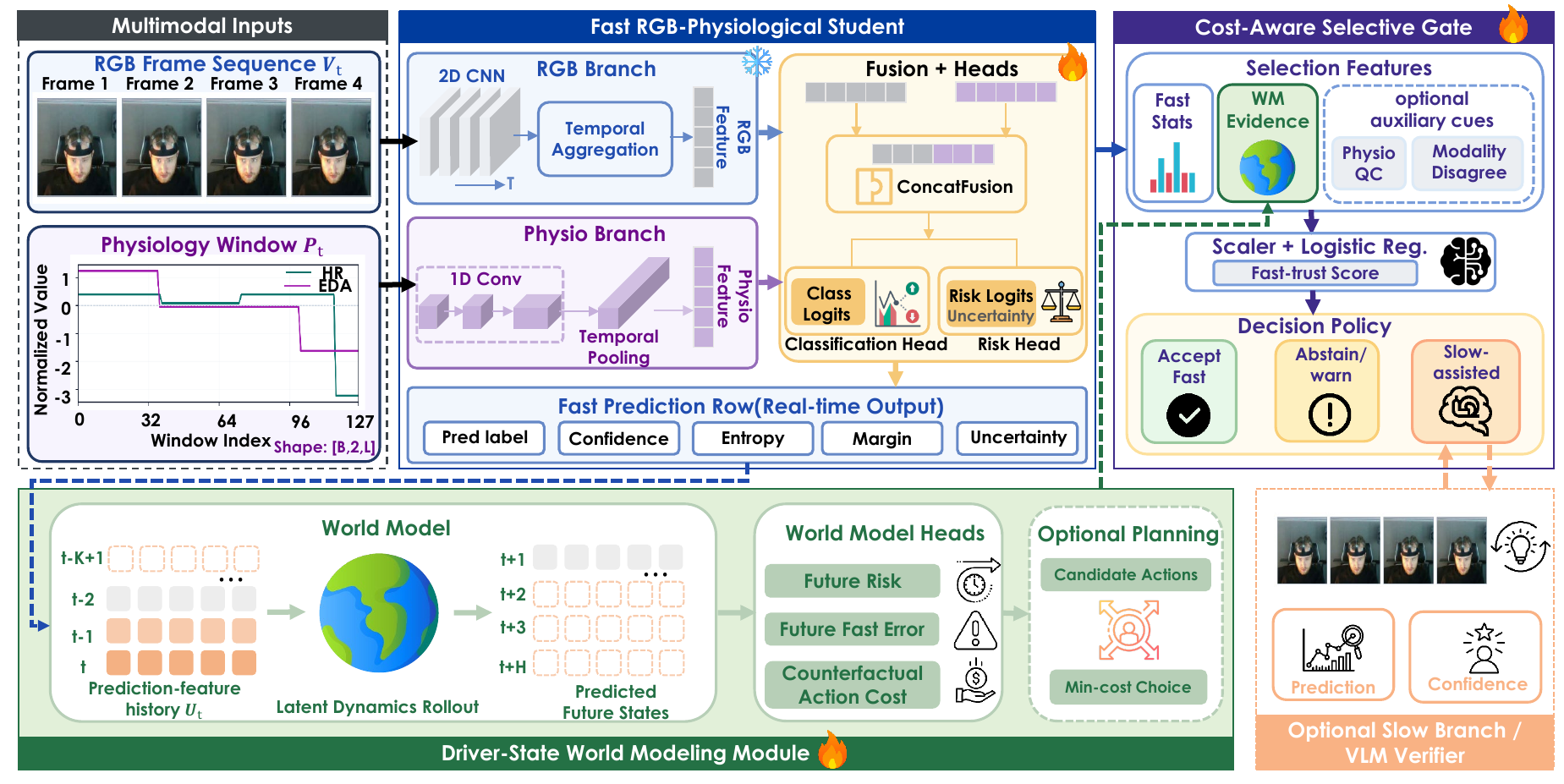} 
    \caption{\textbf{Overview of the proposed framework.}
The system continuously processes RGB frames and window-level HR/EDA signals through a lightweight fast student. Instead of forcing a mandatory classification, a learned cost-aware gate evaluates instantaneous reliability and predictive evidence from a compact driver-state world modeling module. The gate then decides whether to \texttt{accept\_fast}, \texttt{abstain\_warn}, \texttt{slow\_replace}, or \texttt{slow\_verify}. The slow branch is optional and used only for selective replacement or verification.}
    \label{fig:main_architecture}
\end{figure}
% ======================================================

We propose a deployable multimodal driver monitoring framework that separates low-latency perception from risk-aware decision making. The system contains a fast RGB-physiological student, a cost-aware selective gate, and a compact driver-state world modeling module. The student performs continuous monitoring from RGB and physiological signals. The gate decides whether to accept the fast prediction, abstain for safety intervention, or invoke an optional slow branch. The world model supplies predictive evidence by estimating future fast-model errors and counterfactual system-level action outcomes.

\subsection{Problem Formulation and System Overview}
\label{sec:method_overview}

Let $\mathbf{x}_t=(\mathbf{V}_t,\mathbf{P}_t)$ denote the multimodal observation at time window $t$, where $\mathbf{V}_t$ is a short in-cabin RGB clip and $\mathbf{P}_t$ is the synchronized physiological window containing heart-rate and electrodermal activity signals. The target label is $y_t\in\mathcal{Y}$ with $\mathcal{Y}=\{\texttt{low},\texttt{high}\}$, where \texttt{high} is treated as the safety-critical positive state. Unlike a conventional classifier that always outputs a label, our system also predicts a monitoring action
\begin{equation}
\mathcal{A}=\{\texttt{accept\_fast},\texttt{abstain\_warn},\texttt{slow\_replace},\texttt{slow\_verify}\}.
\end{equation}
Here, \texttt{accept\_fast} uses the fast-student prediction, \texttt{abstain\_warn} suppresses the class output and triggers conservative intervention, \texttt{slow\_replace} uses the optional slow-branch prediction, and \texttt{slow\_verify} accepts only when the fast and slow predictions agree. The complete system outputs $(\hat{y}_t,a_t)=\Pi(\mathbf{V}_t,\mathbf{P}_t)$ with $a_t\in\mathcal{A}$. The objective is therefore not only to maximize recognition accuracy, but also to reduce unsafe false negatives under coverage and latency constraints.

\subsection{Fast RGB-Physiological Student}
\label{sec:method_student}

The fast student is a lightweight multimodal classifier for continuous deployment. Given a clip $\mathbf{V}_t=\{\mathbf{I}_{t,1},\ldots,\mathbf{I}_{t,T}\}$, the RGB branch extracts frame-level features $\mathbf{r}_{t,i}=f_{\mathrm{rgb}}(\mathbf{I}_{t,i})$ and aggregates them as $\mathbf{r}_t=g_{\mathrm{temp}}(\mathbf{r}_{t,1},\ldots,\mathbf{r}_{t,T})$, where $g_{\mathrm{temp}}$ can be mean pooling, a recurrent unit, a temporal convolution, or a lightweight attention module. The physiological branch encodes the HR/EDA window $\mathbf{P}_t\in\mathbb{R}^{C\times L}$, $C=2$, as $\mathbf{p}_t=f_{\mathrm{phy}}(\mathbf{P}_t)$. The fused representation is
\begin{equation}
\mathbf{h}_t=f_{\mathrm{fuse}}([\mathbf{r}_t;\mathbf{p}_t]),\qquad
\mathbf{s}_t=W_c\mathbf{h}_t+\mathbf{b}_c,\qquad
\mathbf{q}_t=\mathrm{softmax}(\mathbf{s}_t),
\end{equation}
where $\mathbf{s}_t$ are logits and $\mathbf{q}_t$ is the predicted distribution. The fast prediction is $\hat{y}^{\,f}_t=\arg\max_{y\in\mathcal{Y}}\mathbf{q}_t(y)$. The student is trained with the supervised classification loss
\begin{equation}
\mathcal{L}_{\mathrm{cls}}=-\sum_t \log \mathbf{q}_t(y_t).
\end{equation}
The hidden representation $\mathbf{h}_t$, logits $\mathbf{s}_t$, probabilities $\mathbf{q}_t$, and uncertainty statistics are reused by the selective gate and the driver-state world model.

\subsection{Cost-Aware Selective Inference}
\label{sec:method_selective}

The selective gate estimates whether the fast prediction should be trusted. For each sample, we construct a reliability vector
\begin{equation}
\boldsymbol{\phi}_t=[c_t,e_t,m_t,d_t,q^{\mathrm{phy}}_t,\boldsymbol{\eta}_t],
\end{equation}
where $c_t=\max_y\mathbf{q}_t(y)$ is confidence, $e_t$ is predictive entropy, $m_t$ is the logit margin between the top two classes, $d_t$ is disagreement between RGB-only and RGB-physiological predictions, $q^{\mathrm{phy}}_t$ summarizes physiological signal quality, and $\boldsymbol{\eta}_t$ contains optional predictive features from the driver-state world model. The learned gate is a reliability estimator $\rho_t=g_{\omega}(\boldsymbol{\phi}_t)$, where $\rho_t$ is the estimated probability that the fast decision is correct. A calibrated policy chooses $a_t=\pi_{\omega}(\boldsymbol{\phi}_t)$; in its threshold form,
\begin{equation}
a_t=
\begin{cases}
\texttt{accept\_fast}, & \rho_t\geq\tau,\\
\texttt{fallback}, & \rho_t<\tau,
\end{cases}
\end{equation}
where \texttt{fallback} is instantiated as \texttt{abstain\_warn}, \texttt{slow\_replace}, or \texttt{slow\_verify}. The fallback mode, reliability threshold, and optional slow-confidence floor are selected on calibration data and fixed during testing.

For an action $a\in\mathcal{A}$, let $\tilde{y}_t(a)$ be the final label produced by the action; if the action abstains, $\tilde{y}_t(a)=\varnothing$. We use the action utility
\begin{equation}
\begin{aligned}
U(a;\hat{y}^{\,f}_t,y_t)
=&\;R_{\mathrm{cor}}\mathbf{1}[\tilde{y}_t(a)=y_t]
-C_{\mathrm{ufn}}\mathbf{1}[\tilde{y}_t(a)\neq\varnothing,\,y_t=\texttt{high},\,\tilde{y}_t(a)\neq y_t] \\
&-C_{\mathrm{err}}\mathbf{1}[\tilde{y}_t(a)\neq\varnothing,\,\tilde{y}_t(a)\neq y_t,\,y_t\neq\texttt{high}]\\
&-C_{\mathrm{abs}}\mathbf{1}[\tilde{y}_t(a)=\varnothing] 
-C_{\mathrm{lat}}\ell(a),
\end{aligned}
\end{equation}
where $R_{\mathrm{cor}}$ rewards accepted correct predictions, $C_{\mathrm{ufn}}$ penalizes unsafe false negatives, $C_{\mathrm{err}}$ penalizes other accepted errors, $C_{\mathrm{abs}}$ penalizes abstention, and $\ell(a)$ is the latency of action $a$. The unsafe-false-negative penalty is larger than the ordinary error penalty.

The gate is trained and calibrated in two stages. First, $g_{\omega}$ is trained to predict fast-decision correctness with cost-sensitive sample weights. Second, candidate thresholds and fallback modes are evaluated on a held-out calibration subset, and the operating point is selected by maximizing empirical utility:
\begin{equation}
(\tau^\star,b^\star)=
\arg\max_{\tau,b}
\frac{1}{|\mathcal{D}_{\mathrm{cal}}|}
\sum_{t\in\mathcal{D}_{\mathrm{cal}}}
U(\pi_{\omega}^{\tau,b}(\boldsymbol{\phi}_t);\hat{y}^{\,f}_t,y_t),
\end{equation}
where $b$ denotes the fallback mode. At inference time,
\begin{equation}
\hat{y}_t=
\begin{cases}
\hat{y}^{\,f}_t, & a_t=\texttt{accept\_fast},\\
\hat{y}^{\,s}_t, & a_t=\texttt{slow\_replace},\\
\hat{y}^{\,f}_t, & a_t=\texttt{slow\_verify}\ \text{and}\ \hat{y}^{\,s}_t=\hat{y}^{\,f}_t,\\
\varnothing, & \text{otherwise},
\end{cases}
\end{equation}
where $\hat{y}^{\,s}_t$ is the optional slow-branch prediction. Abstention is interpreted as safety intervention rather than a normal classification decision. We therefore report coverage, unsafe false-negative rate, positive abstention, unhandled positives, slow-call rate, and effective latency in addition to standard classification metrics.

\subsection{Driver-State World Modeling}
\label{sec:method_world_model}

The driver-state world model provides predictive evidence for the gate. Unlike scene-level autonomous-driving world models that generate future external environments, our module operates on compact latent driver-state features. It estimates short-horizon monitoring risk and counterfactual system-level costs rather than replacing the fast student. The design factorizes driver-state dynamics from system-action outcomes: the monitoring action is not assumed to causally change the driver's physiological state within the short decision window. We therefore model natural driver-state evolution with an action-free latent rollout and system-action consequences with a separate action-conditioned branch.

For each window, we define
\begin{equation}
\mathbf{u}_t=[\mathbf{h}_t;\mathbf{s}_t;c_t;e_t;m_t;d_t;q^{\mathrm{phy}}_t],
\end{equation}
where the terms denote the fast-student hidden representation, logits, confidence, entropy, margin, modality disagreement, and physiological quality. In implementation, $\mathbf{u}_t$ may also include reference-branch statistics and additional physiological-quality statistics used by the gate. Given the feature history $\mathbf{U}_t=\{\mathbf{u}_{t-K+1},\ldots,\mathbf{u}_t\}$, the encoder produces the latent state $\mathbf{z}_t=E_{\theta}(\mathbf{U}_t)$.

\paragraph{Action-free driver-state rollout.}
Starting from $\hat{\mathbf{z}}_t=\mathbf{z}_t$, the model performs closed-loop latent rollout by predicting a compact future feature and feeding it back into the next transition:
\begin{equation}
\hat{\mathbf{u}}_{t+k}=G_{\theta}(\hat{\mathbf{z}}_{t+k}),\qquad
\hat{\mathbf{f}}_{t+k}=F_{\theta}(\hat{\mathbf{u}}_{t+k}),\qquad
\hat{\mathbf{z}}_{t+k+1}=\hat{\mathbf{z}}_{t+k}+T_{\theta}(\hat{\mathbf{z}}_{t+k},\hat{\mathbf{f}}_{t+k}).
\end{equation}
This feedback rollout predicts future high-demand probability $\hat{r}_{t+k}=R_{\theta}(\hat{\mathbf{z}}_{t+k})$ and future fast-model error $\hat{e}_{t+k}=E^{\mathrm{err}}_{\theta}(\hat{\mathbf{z}}_{t+k})$. We also use aggregate horizon predictions
\begin{equation}
\hat{r}^{+}_{t}=\max_{1\leq k\leq H}\hat{r}_{t+k},
\qquad
\hat{e}^{+}_{t}=\max_{1\leq k\leq H}\hat{e}_{t+k}.
\end{equation}
No raw RGB frames or physiological waveforms are reconstructed.

\paragraph{Action-conditioned system-outcome estimation.}
For each candidate action $a\in\mathcal{A}$, an action embedding $\mathbf{a}$ is used by a separate outcome branch:
\begin{equation}
\tilde{\mathbf{z}}^a_t=\mathbf{z}_t,\qquad
\tilde{\mathbf{z}}^a_{t+k+1}
=
\tilde{\mathbf{z}}^a_{t+k}
+
T^a_{\theta}(\tilde{\mathbf{z}}^a_{t+k},\tilde{\mathbf{f}}^a_{t+k},\mathbf{a}).
\end{equation}
This branch predicts system-level consequences of taking action $a$ under the current driver-state trajectory; it does not imply that the action changes the driver's natural state. The rollout states are temporally pooled as
\begin{equation}
\bar{\mathbf{z}}^a_t=\operatorname{Pool}(\tilde{\mathbf{z}}^a_{t+1},\ldots,\tilde{\mathbf{z}}^a_{t+H}),
\end{equation}
and action-level outputs are
\begin{equation}
\hat{c}_{t}(a)=C_{\theta}(\bar{\mathbf{z}}^a_t),\qquad
\hat{o}_{t}(a)=O_{\theta}(\bar{\mathbf{z}}^a_t),\qquad
\hat{b}_{t}(a)=B_{\theta}(\bar{\mathbf{z}}^a_t),
\end{equation}

where $\hat{c}_{t}(a)$ is the predicted counterfactual action cost, $\hat{o}_{t}(a)$ estimates unsafe-outcome probability, and $\hat{b}_{t}(a)$ estimates abstention probability. These are action-level summaries over the rollout horizon.

\paragraph{Training objective and gate features.}
Let $r_{t+k}$ and $e_{t+k}$ denote future high-demand and future fast-error targets, and define $r_t^+=\max_{1\leq k\leq H}r_{t+k}$ and $e_t^+=\max_{1\leq k\leq H}e_{t+k}$. The world model is trained with
\begin{equation}
\begin{aligned}
\mathcal{L}_{\mathrm{wm}}
=&\lambda_{r^+}\mathrm{BCE}(\hat{r}^{+}_{t},r^{+}_{t})
+\lambda_{e^+}\mathrm{BCE}(\hat{e}^{+}_{t},e^{+}_{t})
+\lambda_{r_1}\mathrm{BCE}(\hat{r}_{t+1},r_{t+1})\\
&+\sum_{k=1}^{H}
\Big[
\lambda_r\mathrm{BCE}(\hat{r}_{t+k},r_{t+k})
+\lambda_e\mathrm{BCE}(\hat{e}_{t+k},e_{t+k})
+\lambda_u\|\hat{\mathbf{u}}_{t+k}-\mathbf{u}_{t+k}\|_2^2
\Big]\\
&+\lambda_c\ell_c(\hat{\mathbf{c}}_t,\mathbf{c}_t)
+\lambda_o\mathrm{BCE}(\hat{\mathbf{o}}_t,\mathbf{o}_t)
+\lambda_b\mathrm{BCE}(\hat{\mathbf{b}}_t,\mathbf{b}_t),
\end{aligned}
\end{equation}
where $\hat{\mathbf{c}}_t$, $\hat{\mathbf{o}}_t$, and $\hat{\mathbf{b}}_t$ collect predicted costs, unsafe-outcome probabilities, and abstention probabilities for all actions. The targets $\mathbf{c}_t$, $\mathbf{o}_t$, and $\mathbf{b}_t$ are obtained by offline counterfactual evaluation using the ground-truth label, fast prediction, optional slow prediction, and action-specific latency or intervention costs. The feature consistency term aligns predicted compact driver-state features with future observed compact features.

At inference time, the world model augments the gate with
\begin{equation}
\boldsymbol{\eta}_{t}=
\Psi(\hat{r}^{+}_{t},\hat{e}^{+}_{t},\hat{r}_{t+1:t+H},\hat{e}_{t+1:t+H},\hat{\mathbf{c}}_{t},\hat{\mathbf{o}}_{t},\hat{\mathbf{b}}_{t}),
\end{equation}
where $\Psi(\cdot)$ summarizes aggregate risk, selected rollout-step probabilities, counterfactual action costs, unsafe-action probabilities, and model-preferred action summaries. The selective gate uses $\boldsymbol{\eta}_t$ together with instantaneous confidence and modality-reliability signals to choose the final system action.

\section{Experiments}
\label{sec:experiments}

We evaluate the proposed framework on scenario-induced driver-demand monitoring with RGB and physiological signals. The experiments address four questions: whether physiological signals improve the fast student, whether selective inference reduces unsafe accepted errors, whether the system remains deployable under latency constraints, and whether driver-state world modeling provides useful predictive evidence under group shift.

\subsection{Experimental Setup}
\label{sec:exp_setup}

\textbf{Task and inputs.}
The main evaluation is conducted on manD. Each sample contains an in-cabin RGB clip and a window-level physiological segment with HR and EDA signals. The target label is \texttt{low} or \texttt{high}, where \texttt{high} is treated as the safety-critical positive state. The system output is not limited to a class label: the selective policy either accepts the fast prediction, abstains for intervention, or optionally invokes a slow branch.

\textbf{Metrics.}
For classification, we report Macro-F1 and balanced accuracy (BAcc). For selective monitoring, we report coverage, unsafe false-negative rate, positive abstention, unhandled positive rate, slow-call rate, and effective latency. An unsafe false negative is an accepted prediction where a true \texttt{high} sample is output as \texttt{low}. Positive abstention counts true \texttt{high} samples rejected for intervention, and unhandled positive rate is the sum of unsafe false negatives and positive abstentions. Gate thresholds, fallback modes, and operating points are selected on calibration data and fixed for test evaluation.

\subsection{Multimodal Student and Reliability Checks}
\label{sec:exp_student}

Table~\ref{tab:core_results}(a) summarizes the modality ablation and reliability controls. The physiology-only model is weak, while the RGB-only student reaches 0.6608 Macro-F1 and 0.7463 BAcc. Fusing RGB with HR/EDA improves performance to 0.7440 Macro-F1 and 0.9099 BAcc, establishing the RGB-physiological student as the strongest always-on perception model. Shuffling physiological windows substantially degrades performance, indicating that the physiological branch is not a vacuous input. However, shifted physiology remains competitive, and the scenario-only baseline is strong. We therefore interpret the task as scenario-induced driver-demand monitoring rather than clinical stress or workload diagnosis.

We also evaluate architecture choices, physiological normalization, and VLM-to-student distillation. These results are provided in the supplementary material. In brief, ResNet18+GRU gives the best BAcc among tested temporal modules, MobileNetV3-small gives the best Macro-F1, per-window z-score is the most stable physiological normalization, and neither raw nor reliability-filtered KD consistently improves over the supervised RGB-physiological student.
\begin{table}[t]
\centering
\caption{Core results on manD. (a) Modality and reliability controls; $\dagger$ denotes a diagnostic probe, not a deployable monitoring policy. (b) Selective monitoring and deployment. Gate results are mean $\pm$ std across seeds.}
\label{tab:core_results}
\vspace{-1mm}
\resizebox{0.98\linewidth}{!}{%
\begin{tabular}{@{}l cc @{\hspace{3.5em}} l ccc@{}}
\toprule
\multicolumn{3}{c}{\textbf{(a) Modality and reliability}} &
\multicolumn{4}{c}{\textbf{(b) Selective monitoring}} \\
\cmidrule(r{1.5em}){1-3}\cmidrule(l{1.5em}){4-7}
Setting & \textbf{Macro-F1}$_{\uparrow}$ & \textbf{BAcc}$_{\uparrow}$ &
Metric & Fast & Slow & \textbf{Gate (Ours)} \\
\midrule
\multicolumn{3}{@{}l}{\textit{Perception baselines}} & \multicolumn{4}{l}{\textit{Standard classification metrics}} \\
\textcolor{gray}{Majority (High)} & \textcolor{gray}{0.4710} & \textcolor{gray}{0.5000} &
Coverage$_{\uparrow}$ & 1.0000 & 1.0000 & \cellcolor{gray!10}$0.8438{\pm}0.0379$ \\
Physio-only & 0.3607 & 0.5450 &
Slow-call$_{\downarrow}$ & 0.0000 & 1.0000 & \cellcolor{gray!10}$0.0101{\pm}0.0175$ \\
RGB-only & 0.6608 & 0.7463 &
Macro-F1$_{\uparrow}$ & 0.7440 & 0.4506 & \cellcolor{gray!10}\textbf{0.8421}${\pm}0.0214$ \\
\cellcolor{gray!10}\textbf{RGB+physio (Ours)} & \cellcolor{gray!10}\textbf{0.7440} & \cellcolor{gray!10}\textbf{0.9099} &
BAcc$_{\uparrow}$ & 0.9099 & 0.4442 & \cellcolor{gray!10}\textbf{0.9463}${\pm}0.0353$ \\
\midrule
\multicolumn{3}{@{}l}{\textit{Reliability controls}} & \multicolumn{4}{l}{\textit{Safety-critical \& deployment}} \\
Shuffled physio & 0.5987 & 0.8029 &
Unsafe FN$_{\downarrow}$ & 0.1737 & 0.2104 & \cellcolor{gray!10}$\mathbf{0.0527{\pm}0.0081}$ \\
Shifted physio & 0.7432 & 0.8569 &
Pos. abstain & 0.0000 & 0.0000 & \cellcolor{gray!10}$0.1272{\pm}0.0335$ \\
Scenario-only & 0.7372 & 0.9089 &
Unhandled$_{\downarrow}$ & 0.1737 & 0.2104 & \cellcolor{gray!10}$0.1799{\pm}0.0254$ \\
\textit{Scenario+score probe}$^\dagger$ & \textbf{0.7803} & \textbf{0.9239} &
Latency$_{\downarrow}$ & 3.08 ms & 8615.98 ms & \cellcolor{gray!10}90.18 ms \\
\bottomrule
\end{tabular}%
}
\vspace{-2mm}
\end{table}

\paragraph{Beyond scenario priors.}
Because scenario priors are strong in manD, we further test whether the learned scores contain information beyond scenario identity. A scenario-only probe obtains 0.7372 Macro-F1 and 0.9089 BAcc, while adding model scores improves performance to 0.7803 Macro-F1 and 0.9239 BAcc. This indicates that the predictions are not reducible to scenario lookup. We also find that the RGB-physiological student recovers 382 of 803 RGB-only errors (47.6\%) and 228 of 646 RGB-only unsafe false negatives (35.3\%), showing that physiological-window evidence is especially useful on visually ambiguous or unsafe RGB failures. A stricter test-time physiology intervention study in the supplement shows that the current model is not strongly sensitive to exact physiological synchronization; we therefore interpret HR/EDA as window-level contextual evidence rather than strictly time-locked physiological dynamics.

\subsection{Cost-Aware Selective Inference and Deployment}
\label{sec:exp_gate}

Table~\ref{tab:core_results}(b) reports the core selective-inference and deployment results. Always-fast inference accepts every sample and produces an unsafe false-negative rate of 0.1737. The learned cost-aware gate reduces this to $0.0527\pm0.0081$ across seeds while improving Macro-F1 and BAcc. The learned policy primarily selects fast abstention rather than frequent slow escalation, which is consistent with the large latency gap between the fast student and the slow VLM. The fast student contains 11.39M parameters and runs in 3.08 ms, while the slow VLM requires 8615.98 ms per sample and is unsuitable as a continuous monitor.

\begin{figure}[t]
    \centering
    \includegraphics[width=0.8\linewidth]{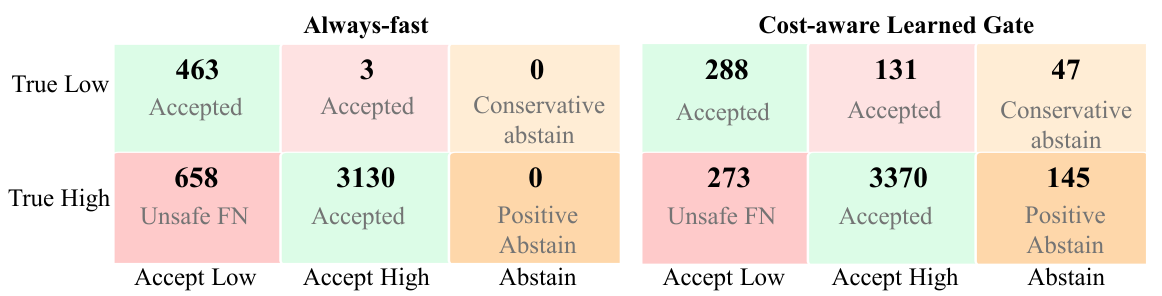}
    \caption{\textbf{Selective confusion matrix.} Comparison between always-fast inference and the learned cost-aware gate. By explicitly optimizing for asymmetric risk, the learned gate successfully redistributes safety-critical errors (Unsafe FNs, red) into conservative or positive abstentions (orange), while simultaneously increasing the number of correctly accepted high-demand states.}
    \label{fig:selective_confusion}
    
\end{figure}

\begin{figure}[t]
    \centering
    \includegraphics[width=\linewidth]{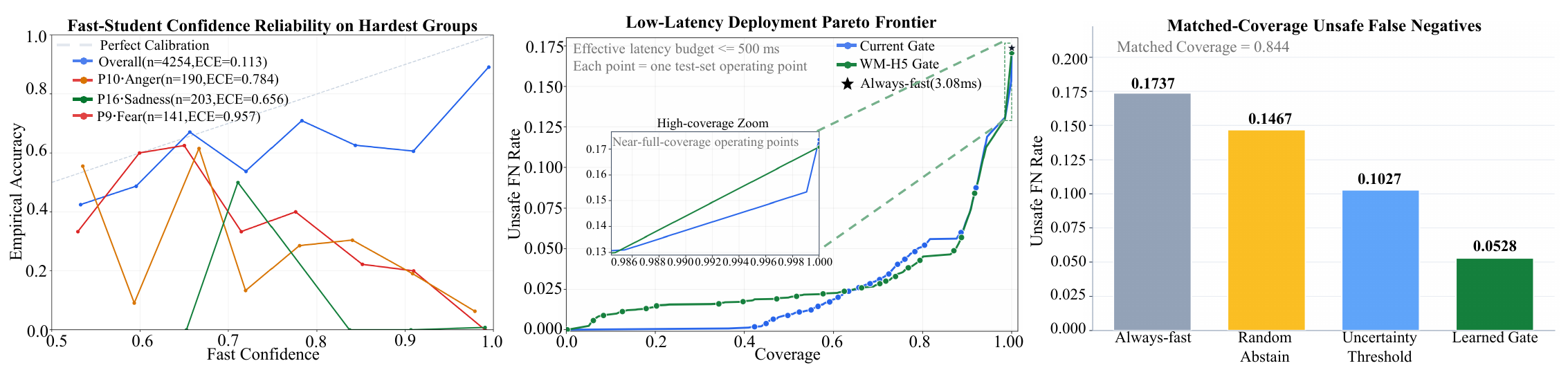}
    \caption{\textbf{Calibration, deployment frontier, and matched-coverage safety behavior.}
    \textbf{Left:} The fast student is reasonably calibrated overall, but reliability degrades markedly on the hardest participant-scenario groups, revealing calibration drift under heavy-tailed group shift.
    \textbf{Middle:} Coverage--risk Pareto frontiers under deployment latency constraints. The predictive driver-state gate improves the trade-off in part of the high-coverage regime, while always-fast inference defines the latency floor.
    \textbf{Right:} Unsafe false-negative rate at matched coverage. The learned gate outperforms random abstention and uncertainty-threshold baselines at the same operating coverage, showing that its safety gain is not explained by lower coverage alone.}
    \label{fig:risk_drift}
\end{figure}

The selective confusion matrix and coverage-risk curves are shown in Fig.~\ref{fig:selective_confusion} and Fig.~\ref{fig:risk_drift}. These visualizations make explicit how cost-aware selection transfers part of the high-risk region from accepted false negatives to abstention or safety intervention.

\paragraph{Matched-coverage comparison.}
At the learned gate's coverage of 0.8439, random abstention yields an unsafe false-negative rate of 0.1467, and confidence, entropy, or margin thresholding yields 0.1027. The learned gate obtains 0.0528 at the same coverage. Thus, the reduction in unsafe accepted errors is not explained by lower coverage alone; the learned gate identifies a more informative rejection region. We additionally report low-risk abstention and warning precision in the supplement, ensuring that abstention is evaluated as an intervention burden rather than treated as a free correct prediction.

\subsection{Driver-State World Modeling}
\label{sec:exp_world_model}

Table~\ref{tab:main_world_model} summarizes the driver-state world-model results. The action-feedback model provides non-trivial predictive evidence: future fast-error AUC increases from 0.8818 at $H=1$ to 0.9486 at $H=5$, and counterfactual action-cost diagnostics improve after action-feedback modeling; detailed horizon-specific cost results are provided in the supplement. When integrated into the gate, the world-model gate with $H=5$ reduces accepted-risk in grouped evaluation, lowering unsafe false negatives from 0.0721 to 0.0507 and improving utility. However, it slightly increases unhandled positives. Factorized variants shift the trade-off further: the latent variant is more safety-oriented but abstains more, while the no-latent variant improves classification quality but increases unhandled positives. Thus, driver-state world modeling is a useful predictive extension, but not a uniformly dominant controller.

\begin{table*}[t]
\centering
\small
\caption{Driver-state world modeling. The world-model gate with horizon $H=5$ provides useful predictive evidence, but does not dominate all safety metrics.}
\label{tab:main_world_model}
\resizebox{\linewidth}{!}{%
\begin{tabular}{@{}l c cc cccc c@{}}
\toprule
\textbf{Method} & 
\textbf{Fast-Err AUC}$_{\uparrow}$ & 
\textbf{Cost MSE}$_{\downarrow}$ & \textbf{Pairwise}$_{\uparrow}$ & 
\textbf{Coverage}$_{\uparrow}$ & 
\textbf{Macro-F1}$_{\uparrow}$ & 
\textbf{BAcc}$_{\uparrow}$ & 
\textbf{Unsafe FN}$_{\downarrow}$ & 
\textbf{Unhand. Pos.}$_{\downarrow}$ \\
\midrule
\multicolumn{9}{@{}l}{\textit{Instantaneous baseline}} \\
Current-evidence gate & -- & -- & -- & 0.9549 & 0.7656 & 0.8062 & 0.0721 & \textbf{0.1103} \\
\midrule
\multicolumn{9}{@{}l}{\textit{Predictive extensions (Ours)}} \\
World-model gate ($H=5$) & 0.9486 & 2.2440 & 0.6611 & 0.9339 & 0.8014 & 0.8219 & 0.0510 & 0.1140 \\
\rowcolor{gray!10}
\textbf{Risk-calib. WM gate ($H=5$)} & 0.9486 & 2.2440 & 0.6611 & 0.9342 & 0.8018 & 0.8220 & \textbf{0.0507} & 0.1135 \\
Factorized WM (latent) & -- & -- & -- & 0.8766 & 0.8340 & 0.8954 & 0.0515 & 0.1595 \\
Factorized WM (no latent) & -- & -- & -- & 0.8260 & \textbf{0.8538} & \textbf{0.9634} & 0.0536 & 0.2043 \\
\bottomrule
\end{tabular}%
}
\end{table*}

\begin{figure}[t]
    \centering
    % ---------- 左侧图：多模态对比 ----------
    \begin{minipage}[t]{0.45\linewidth}
        \centering
        \includegraphics[width=\linewidth]{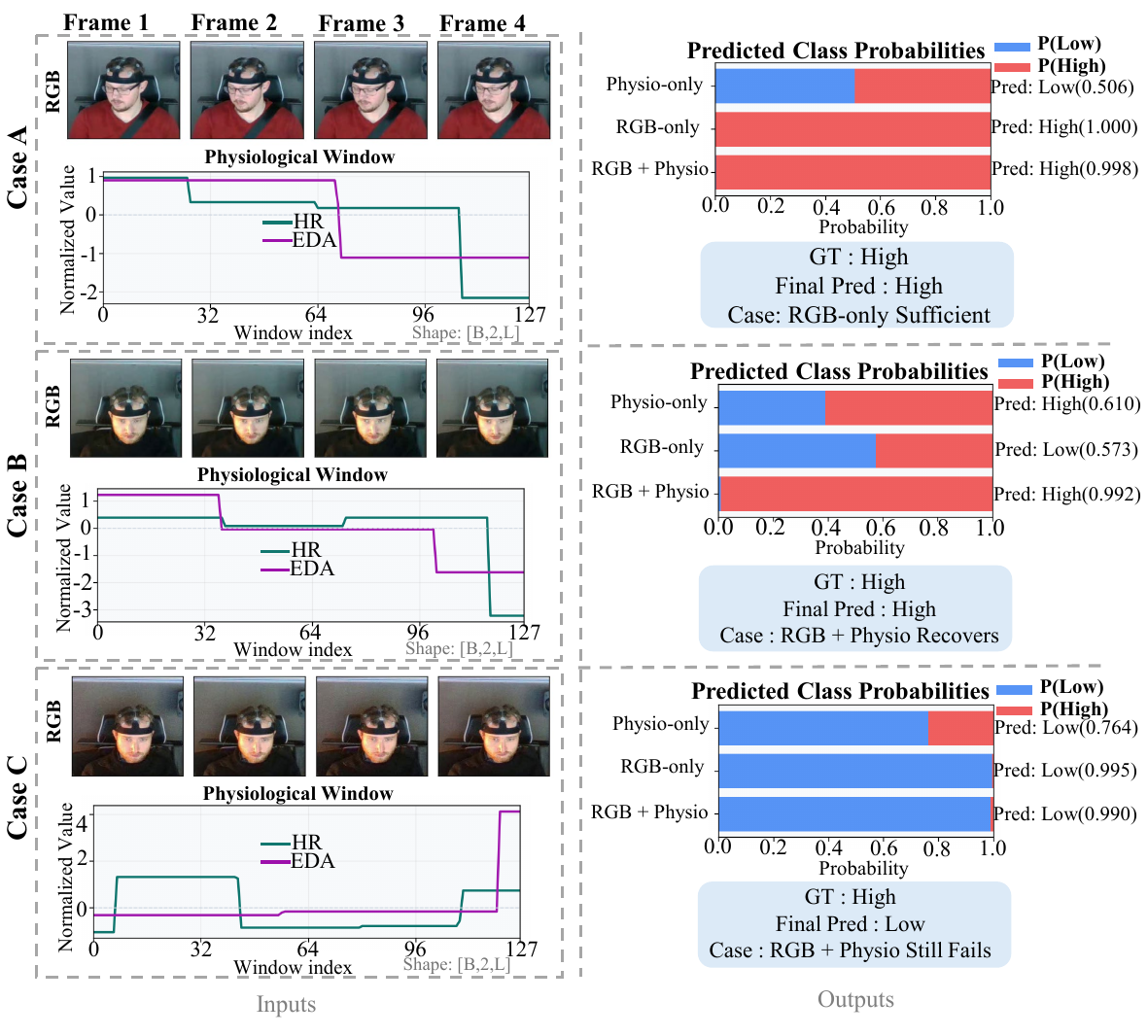}
        \caption{\textbf{Qualitative comparison of modality contributions.} Case A shows visual cues are sufficient. Case B demonstrates the necessity of multimodal fusion: when visual features are ambiguous (head down), physiological dynamics (EDA/HR drop) successfully recover the true \texttt{High} demand state. Case C shows a failure case where both modalities fail to capture the state change.}
        \label{fig:qualitative_modality}
    \end{minipage}
    \hfill % 撑开中间的间距
    % ---------- 右侧图：世界模型推演 ----------
    \begin{minipage}[t]{0.53\linewidth}
        \centering
        \includegraphics[width=\linewidth]{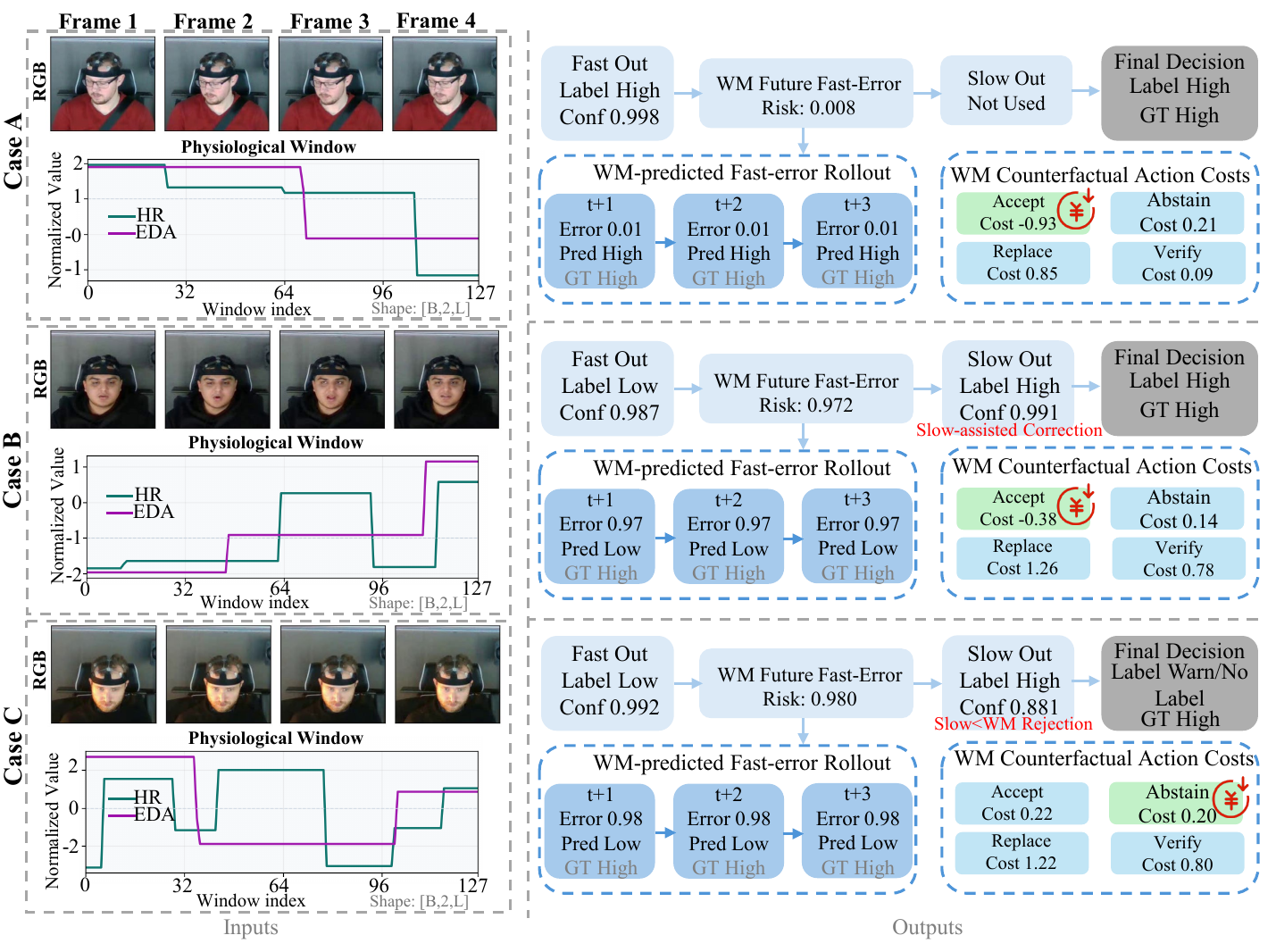}
        \caption{\textbf{Driver-state world model rollout and action-cost estimation.} The world model provides predictive evidence by simulating future per-step risks and counterfactual system-level action costs. In Case B and C, although the instantaneous fast output is incorrect (\texttt{Low}), the world model anticipates high future risk and estimates that \texttt{Abstain} or \texttt{Replace} yields the lowest cost, thereby guiding the gate toward a safer intervention.}
        \label{fig:wm_rollout}
    \end{minipage}    
\end{figure}

The world-model rollout visualization is shown in Fig.~\ref{fig:wm_rollout}. It illustrates how future fast-error estimates and counterfactual action costs are converted into predictive features for the gate.

\subsection{Calibration Drift and Worst-Group Robustness}
\label{sec:exp_robustness}

We further evaluate operating-point stability under participant-scenario shift. In the strict worst-group evaluation, the current-evidence gate obtains lower weighted unsafe FN and unhandled-positive rates than the horizon-$5$ world-model gate, while the world-model gate gives slightly higher utility. Validation-selected fixed-risk operating points also drift on test groups, indicating that predictive evidence alone does not solve group-dependent calibration. The hardest groups are \texttt{P10::Anger}, \texttt{P16::Sadness}, and \texttt{P9::Fear}. Detailed worst-group and fixed-risk results are provided in the supplement.

Together with the matched-coverage comparison in Sec.~\ref{sec:exp_gate}, these results suggest that the learned gate captures useful risk structure on average, but maintaining a fixed operating point under participant-scenario shift remains unresolved.

\subsection{Qualitative Analysis}
\label{sec:exp_qualitative}

We provide qualitative visualizations to illustrate system behavior. Fig.~\ref{fig:qualitative_modality} compares RGB-only and RGB-physiological predictions using input waveforms and prediction distributions. Fig.~\ref{fig:selective_confusion} shows how selective inference separates accepted \texttt{low}, accepted \texttt{high}, and abstained samples. Fig.~\ref{fig:wm_rollout} visualizes driver-state rollout predictions and counterfactual action costs. Finally, Fig.~\ref{fig:risk_drift} summarizes the overarching deployment characteristics, illustrating both the optimal coverage-risk Pareto frontier and the persistent calibration drift observed on difficult participant-scenario groups.

\paragraph{Supplementary results.}
The supplementary material includes VLM distillation, student architecture and normalization ablations, per-seed learned-gate results, rule-based gate operating points, early and factorized world-model variants, calibration follow-ups, leave-one-scenario-out stress analysis, and synthetic transition forecasting.

\section{Conclusion}
\label{sec:conclusion}

We presented a deployable multimodal driver monitoring framework that shifts in-cabin driver-state recognition from always-accept classification to cost-aware selective inference. The system combines an always-on RGB-physiological student with a learned gate that accepts reliable fast predictions and abstains under high risk. Our results show that window-level physiological signals help recover visually ambiguous RGB errors and unsafe false negatives, while selective inference further reduces unsafe accepted errors against matched-coverage baselines without sacrificing deployment-level latency. Scenario-prior and intervention analyses also show that exact physiological synchronization remains a limitation, pointing to the need for more robust multimodal alignment.

Our study further clarifies the role of heavier predictive components. Slow VLMs and direct VLM-to-student distillation do not consistently improve the supervised fast student, and should not be assumed to be reliable always-on monitors. Driver-state world modeling provides useful future-risk and action-cost signals, but remains sensitive to operating-point calibration drift under worst-group shifts. These findings suggest that reliable edge driver monitoring requires not only stronger perception backbones, but also risk-aware selection, abstention-aware evaluation, and group-robust calibration.

% \clearpage  % TODO REVIEW/FINAL: This \clearpage needs to be removed from both review and camera-ready versions.

% 仅在 Camera-Ready 阶段取消注释并加入这段代码
\section*{Acknowledgements}
This work was supported by the National Natural Science Foundation of China (Grant No. 62273135), the Hubei Provincial Key Research and Development Program (No. 2025BAB061), the Hubei Provincial Major Science and Technology Program (No. 2025BEA001), and the Wuhan Natural Science Foundation Focused Program (No. 2026040101030022).

% TODO REVIEW/FINAL: This \clearpage needs to be removed from both review and camera-ready versions.
\bibliographystyle{splncs04}
\bibliography{main}

\clearpage
\appendix  % 切换章节编号格式为 A, B, C...
% 建议使用 \input 而不是 \include，因为 \input 更灵活且不会强制分页
\section{Additional Discussion}
\label{supp:discussion}

This section clarifies several design choices that are central to interpreting the proposed framework.

\noindent\textbf{Q1: \textit{Why formulate driver monitoring as selective inference rather than simply maximizing classification accuracy?}}

\smallskip
Continuous driver monitoring is not only a recognition problem. In a deployed vehicle, the system must also decide whether its own prediction is reliable enough to act on. This distinction matters because the error costs are asymmetric. Accepting a wrong \texttt{low} prediction for a true \texttt{high} driver state is more safety-critical than abstaining and triggering a conservative intervention. A standard classifier is forced to assign a label to every input, even when the visual and physiological evidence is ambiguous. Selective inference gives the monitor an explicit reject option: it can accept the fast prediction when the evidence is reliable, and abstain when the risk of an unsafe accepted error is too high. The objective is therefore not only to improve average recognition metrics, but to control the operating behavior of the system under uncertainty.

\medskip
\noindent\textbf{Q2: \textit{Why focus on scenario-induced driver demand instead of visible actions such as phone usage or eating?}}

\smallskip
Visible secondary activities are important, but they do not cover all safety-relevant driver states. A driver may look visually normal while experiencing elevated demand due to the driving scenario, automation state, or internal physiological response. Such demand-related states are only partially observable from RGB frames. This motivates the use of window-level physiological signals, which can provide complementary contextual evidence when visual cues are ambiguous. We therefore study scenario-induced driver-demand monitoring rather than general action recognition. This choice is not meant to replace visible behavior monitoring; it targets a different failure mode, where the driver state may be visually ambiguous but physiologically informative.

\medskip
\noindent\textbf{Q3: \textit{How should the strong scenario-only baseline be interpreted?}}

\smallskip
The scenario-only baseline confirms that the constructed demand labels contain scenario-level structure. We therefore do not interpret the task as clinical stress diagnosis or as a direct measurement of cognitive workload. At the same time, physiological signals are not redundant. The RGB-physiological student improves over RGB-only and physiology-only baselines, and shuffled physiological windows degrade performance. These results indicate that HR/EDA provide useful window-level state information, although part of the signal is correlated with scenario structure. This is why we describe the task as scenario-induced demand monitoring and include shuffled, shifted, and scenario-only controls rather than claiming pure physiological causality.

\medskip
\noindent\textbf{Q4: \textit{Why not rely on a large VLM as the continuous monitor?}}

\smallskip
Large VLMs provide broad semantic priors, but they are poorly matched to continuous edge monitoring. Their autoregressive decoding and large parameter counts lead to inference latency that is incompatible with always-on in-cabin monitoring. More importantly, in our experiments the slow VLM branch is not consistently stronger than the supervised RGB-physiological student, either as a direct predictor or as a distillation teacher. We therefore treat the VLM as an optional probe or rare verification branch, not as the main deployed monitor. The common path must remain lightweight and reliable; uncertainty is handled by selective decision-making rather than by continuously running a large model.

\medskip
\noindent\textbf{Q5: \textit{What is meant by driver-state world modeling in this work?}}

\smallskip
Our driver-state world modeling module is not a scene-level autonomous-driving world model. It does not generate future road images, traffic scenes, occupancy maps, or driver videos. Instead, it operates on compact in-cabin monitoring features extracted from the fast student, including latent representations, logits, uncertainty statistics, modality disagreement, and physiological quality. The module rolls out these driver-state features over a short horizon and estimates future fast-model errors and counterfactual system-level action costs. Its role is to provide predictive evidence to the selective gate, not to replace the fast student or to serve as a standalone simulator.

\medskip
\noindent\textbf{Q6: \textit{Why factorize driver-state dynamics from action-conditioned outcomes?}}

\smallskip
The monitoring action is not assumed to causally change the driver's physiological state within the short decision window. For example, accepting a fast prediction or abstaining for intervention does not provide observed evidence that the driver's HR/EDA dynamics would immediately change in the offline data. Modeling the action as directly controlling driver-state evolution would therefore introduce an unsupported causal assumption. We instead use a factorized formulation: an action-free rollout models the natural evolution of driver-state features, while a separate action-conditioned branch estimates the system-level consequences of candidate actions, such as counterfactual action cost, unsafe outcome probability, and abstention outcome probability. This matches the available supervision and keeps the model's causal interpretation conservative.

\medskip
\noindent\textbf{Q7: \textit{Why is world modeling treated as an extension rather than the main controller?}}

\smallskip
The world model provides useful predictive signals. It improves future fast-error prediction and can reduce accepted risk at some operating points. However, it does not uniformly dominate the current-evidence gate under worst-group evaluation. In particular, reductions in unsafe accepted false negatives can be accompanied by higher abstention or unhandled-positive rates, and operating points selected on validation data may drift under participant-scenario shifts. We therefore use driver-state world modeling as a predictive extension for selective inference rather than as an independent controller. This interpretation reflects the empirical evidence: predictive state modeling is valuable, but robust calibration remains necessary before such signals can safely govern decisions on their own.

\medskip
\noindent\textbf{Q8: \textit{Why evaluate the full system primarily on manD?}}

\smallskip
The full framework requires synchronized in-cabin RGB, continuous physiological signals, and driver-state labels. Many public driver monitoring datasets provide visual behavior labels, but do not include aligned HR/EDA streams for the same monitoring windows. Evaluating those datasets would require removing the physiological branch and changing the safety objective, which would no longer test the full system studied in this work. We therefore evaluate the complete RGB-physiological selective monitoring pipeline on manD, and use multiple internal reliability checks---including shuffled and shifted physiology, scenario-only baselines, worst-group evaluation, and calibration analyses---to probe shortcut behavior and robustness. Broader validation on future datasets with synchronized visual and physiological driver-state labels remains an important direction.

\section{Supplementary Material}
\label{sec:supp}

This supplementary material provides additional implementation details and ablation results omitted from the main paper due to space constraints. We include VLM distillation results, student architecture and normalization ablations, per-seed selective-gate results, rule-based operating points, additional world-model variants, calibration follow-ups, and stress-test analyses.

\subsection{Additional Protocol Details}
\label{sec:supp_protocol}

\textbf{Task.}
The primary task is scenario-induced driver-demand monitoring on manD. Each sample contains an RGB clip and a synchronized HR/EDA physiological window. The semantic label is \texttt{low} or \texttt{high}, where \texttt{high} is treated as the safety-critical positive state.

\textbf{Selective monitoring actions.}
The system action space is
\[
\mathcal{A}=\{\texttt{accept\_fast},\texttt{abstain\_warn},\texttt{slow\_replace},\texttt{slow\_verify}\}.
\]
The always-on path is the RGB-physiological fast student. The slow branch is used only as an optional verification or replacement path. For selective evaluation, unsafe false negatives, positive abstention, unhandled positives, coverage, slow-call rate, and effective latency are computed using the definitions in the main paper.

\textbf{Calibration.}
Gate thresholds, fallback modes, and slow-confidence floors are selected on calibration data and fixed for test evaluation. World-model checkpoints are selected by validation gate utility rather than by standalone prediction quality.

\subsection{VLM-to-Student Distillation}
\label{sec:supp_kd}

Table~\ref{tab:supp_kd} reports the VLM-to-student distillation ablation. Raw KD gives a marginal Macro-F1 gain over CE-only training but substantially reduces BAcc. Reliability-filtered KD is also weaker than the supervised RGB-physiological student. This supports the main-paper conclusion that the VLM teacher is not a consistently stronger training signal in this protocol.

\begin{table}[t]
\centering
\small
\caption{VLM-to-student distillation ablation.}
\label{tab:supp_kd}
\begin{tabular}{lcc}
\toprule
Training strategy & Macro-F1 & BAcc \\
\midrule
CE-only RGB+physio & 0.7440 & \textbf{0.9099} \\
Raw KD & \textbf{0.7461} & 0.7768 \\
Reliability-filtered KD & 0.6828 & 0.8309 \\
\bottomrule
\end{tabular}
\end{table}

\subsection{Student Architecture and Physiological Normalization}
\label{sec:supp_student_ablation}

Table~\ref{tab:supp_arch_norm} reports the student architecture and physiological normalization ablations. Pretrained visual encoders consistently improve the always-on student. Among the tested temporal modules, ResNet18 with GRU aggregation gives the highest BAcc, while MobileNetV3-small gives the highest Macro-F1. For physiological preprocessing, per-window z-score normalization remains the most stable setting overall.

\begin{table}[t]
\centering
\small
\caption{Student architecture and physiological normalization ablations.}
\label{tab:supp_arch_norm}
\resizebox{\linewidth}{!}{
\begin{tabular}{llcc}
\toprule
Group & Variant & Macro-F1 & BAcc \\
\midrule
Main student & RGB+physio & 0.7440 & 0.9099 \\
\midrule
Backbone / temporal & ResNet18 + pretrained + mean & 0.7352 & 0.9071 \\
Backbone / temporal & MobileNetV3-small + pretrained + mean & \textbf{0.7774} & 0.8960 \\
Backbone / temporal & ResNet18 + pretrained + GRU & 0.7674 & \textbf{0.9221} \\
Backbone / temporal & ResNet18 + pretrained + attention & 0.6407 & 0.8508 \\
Backbone / temporal & ResNet18 + pretrained + TCN & 0.6901 & 0.8822 \\
\midrule
Physio normalization & Per-window z-score & 0.7440 & \textbf{0.9099} \\
Physio normalization & Global z-score & \textbf{0.7651} & 0.8677 \\
Physio normalization & Raw statistics & 0.6058 & 0.6659 \\
\bottomrule
\end{tabular}
}
\end{table}

\subsection{Learned Gate Across Seeds}
\label{sec:supp_gate_seeds}

Table~\ref{tab:supp_gate_seeds} reports per-seed results for the learned cost-aware gate. Most seeds select a fast-abstain operating mode rather than frequent slow escalation, which is consistent with the large latency gap between the fast student and the slow branch.

\begin{table*}[t]
\centering
\small
\caption{Learned cost-aware gate across seeds.}
\label{tab:supp_gate_seeds}
\resizebox{\linewidth}{!}{
\begin{tabular}{lcccccccc}
\toprule
Seed & Coverage & Slow-call & Macro-F1 & BAcc & Unsafe FN & Pos. abstain & Unhandled pos. & Latency \\
\midrule
A & 0.8439 & 0.0000 & 0.8498 & 0.9644 & 0.0528 & 0.1259 & 0.1787 & 3.08 ms \\
23 & 0.8373 & 0.0000 & 0.8509 & 0.9650 & 0.0512 & 0.1317 & 0.1829 & 3.08 ms \\
37 & 0.7936 & 0.0000 & \textbf{0.8616} & \textbf{0.9705} & \textbf{0.0420} & 0.1729 & 0.2149 & 3.08 ms \\
51 & \textbf{0.9003} & 0.0404 & 0.8059 & 0.8853 & 0.0647 & \textbf{0.0784} & \textbf{0.1431} & 351.45 ms \\
\midrule
Mean $\pm$ std & $0.8438{\pm}0.0379$ & $0.0101{\pm}0.0175$ & $0.8421{\pm}0.0214$ & $0.9463{\pm}0.0353$ & $0.0527{\pm}0.0081$ & $0.1272{\pm}0.0335$ & $0.1799{\pm}0.0254$ & 90.18 ms \\
\bottomrule
\end{tabular}
}
\end{table*}

\subsection{Rule-Based Selective Operating Points}
\label{sec:supp_rule_gates}

Table~\ref{tab:supp_rule_gates} provides representative rule-based selective operating points. These rules demonstrate the expected risk-coverage-latency trade-off, but they are less stable and less compact than the learned cost-aware gate reported in the main paper.

\begin{table}[t]
\centering
\small
\caption{Representative rule-based selective operating points.}
\label{tab:supp_rule_gates}
\resizebox{\linewidth}{!}{
\begin{tabular}{lccc}
\toprule
Policy & Coverage & Unsafe FN & Effective latency \\
\midrule
Always-fast & 1.0000 & 0.1737 & 3.08 ms \\
Confidence-margin & 0.8902 & 0.0636 & 1451.23 ms \\
Confidence-margin strict & 0.7494 & 0.0449 & 961.09 ms \\
Entropy & 0.6946 & \textbf{0.0314} & 1108.94 ms \\
Physio-quality confidence gate & 0.8839 & 0.0623 & 1505.92 ms \\
Strict physio-quality confidence gate & 0.8467 & 0.0444 & 1880.62 ms \\
Best physio-aware gate & 0.8310 & 0.0335 & 2046.70 ms \\
\bottomrule
\end{tabular}
}
\end{table}

\subsection{Early Driver-State World-Model Variants}
\label{sec:supp_early_wm}

Table~\ref{tab:supp_early_wm} reports early driver-state world-model variants. The GRU and rollout variants provide strong future high-demand prediction, while the MLP and rollout variants show competitive fast-error prediction. These variants confirm that predictive state modeling is not empty, but they do not uniformly improve selective-gate performance.

\begin{table*}[t]
\centering
\small
\caption{Early driver-state world-model variants.}
\label{tab:supp_early_wm}
\resizebox{\linewidth}{!}{
\begin{tabular}{lcccccc}
\toprule
Variant & Future high F1/AUC & Fast-error F1/AUC & Coverage & Macro-F1 & BAcc & Unsafe FN \\
\midrule
Current learned gate & -- & -- & 0.8439 & 0.8498 & 0.9644 & 0.0528 \\
GRU state forecaster & 0.8974 / 0.9628 & 0.6703 / 0.9261 & 0.8639 & 0.8232 & 0.9562 & 0.0692 \\
Current-feature MLP & 0.8947 / 0.9523 & 0.5852 / 0.9036 & 0.8357 & \textbf{0.8581} & \textbf{0.9665} & \textbf{0.0486} \\
Latent rollout forecaster & \textbf{0.9028} / 0.9476 & 0.6257 / 0.8958 & 0.8507 & 0.8452 & 0.9618 & 0.0581 \\
\bottomrule
\end{tabular}
}
\end{table*}

\subsection{Action-Feedback World-Model Horizon Ablation}
\label{sec:supp_wm_horizon}

Table~\ref{tab:supp_wm_horizon} reports horizon-level prediction and counterfactual action evaluation for the action-feedback world model. The best future fast-error prediction is obtained at horizon $H=5$, whereas the lowest action-cost MSE occurs at $H=3$. This motivates reporting predictive-risk and action-cost diagnostics separately.

\begin{table}[t]
\centering
\small
\caption{Action-feedback world-model horizon ablation.}
\label{tab:supp_wm_horizon}
\resizebox{\linewidth}{!}{
\begin{tabular}{lcccccc}
\toprule
Horizon & Future high F1 & Future high AUC & Fast-error F1 & Fast-error AUC & Cost MSE & Pairwise \\
\midrule
$H=1$ & 0.8952 & 0.9544 & 0.5645 & 0.8818 & 2.2481 & \textbf{0.6665} \\
$H=3$ & \textbf{0.8980} & \textbf{0.9549} & 0.6510 & 0.9255 & \textbf{2.2440} & 0.6611 \\
$H=5$ & 0.8972 & 0.9521 & \textbf{0.7114} & \textbf{0.9486} & 2.4428 & 0.6627 \\
\bottomrule
\end{tabular}
}
\end{table}

\subsection{Factorized World Models and MPC-Style Gates}
\label{sec:supp_factorized_mpc}

The round-2 world-model experiments evaluate factorized driver dynamics, latent consistency ablations, and policy-library MPC gates. Table~\ref{tab:supp_factorized} summarizes the factorized variants already discussed in the main paper. The latent factorized model is more conservative, while the no-latent variant improves classification metrics but produces a larger unhandled-positive burden.

\begin{table}[t]
\centering
\small
\caption{Factorized driver-state world-model variants.}
\label{tab:supp_factorized}
\resizebox{\linewidth}{!}{
\begin{tabular}{lccccc}
\toprule
Variant & Coverage & Macro-F1 & BAcc & Unsafe FN & Unhandled pos. \\
\midrule
World-model gate ($H=5$) & 0.9339 & 0.8014 & 0.8219 & \textbf{0.0510} & \textbf{0.1140} \\
Factorized world model (latent) & 0.8766 & 0.8340 & 0.8954 & 0.0515 & 0.1595 \\
Factorized world model (no latent) & 0.8260 & \textbf{0.8538} & \textbf{0.9634} & 0.0536 & 0.2043 \\
\bottomrule
\end{tabular}
}
\end{table}

The policy-library MPC gates are weaker than the learned cost-aware gate in the evaluated setting. Across six seed/system combinations, MPC-style gates obtain coverage in the range 0.8590--0.9006, unsafe false-negative rates in the range 0.1195--0.1414, and unhandled positive rates in the range 0.2356--0.2711. These results suggest that explicit policy-library planning is not yet competitive with the learned selective gate.

\subsection{Calibration Follow-Ups}
\label{sec:supp_calibration}

Table~\ref{tab:supp_scenario_calibration} reports the scenario-stratified calibration follow-up. Naive scenario-specific thresholding does not solve operating-point drift and can degrade the test trade-off. The best reported world-model setting remains the global risk-calibrated horizon-$5$ gate.

\begin{table}[t]
\centering
\small
\caption{Scenario-stratified calibration follow-up.}
\label{tab:supp_scenario_calibration}
\resizebox{\linewidth}{!}{
\begin{tabular}{lcccc}
\toprule
Method & Coverage & Macro-F1 & BAcc & Unsafe FN / Unhandled pos. \\
\midrule
Current gate, risk-calibrated & 0.9549 & 0.7648 & 0.8050 & 0.0721 / 0.1103 \\
Current gate, scenario-calibrated & 0.9462 & 0.7445 & 0.7824 & 0.0755 / 0.1212 \\
World-model gate ($H=5$), risk-calibrated & 0.9342 & \textbf{0.8018} & \textbf{0.8220} & \textbf{0.0507} / \textbf{0.1135} \\
World-model gate ($H=5$), scenario-calibrated & 0.9382 & 0.7750 & 0.8091 & 0.0657 / 0.1238 \\
\bottomrule
\end{tabular}
}
\end{table}

\subsection{Leave-One-Scenario-Out Stress Analysis}
\label{sec:supp_loso}

The leave-one-scenario-out stress test gives 0.4339 weighted Macro-F1, 0.7504 weighted BAcc, and 0.7908 weighted accuracy. Because several held-out scenarios contain single-class test labels, this evaluation should be interpreted as shortcut and stress analysis rather than as a standard generalization benchmark. The most difficult participant-scenario groups are \texttt{P10::Anger}, \texttt{P16::Sadness}, and \texttt{P9::Fear}.

\subsection{Synthetic Transition and Early-Warning Stress Test}
\label{sec:supp_transition}

The original manD split does not support a stable real lead-time claim because most high-demand onsets occur near sequence starts. We therefore construct a synthetic low-to-high transition stress test. With the default best-F1 threshold, all evaluated world-model variants degenerate into near-always-warning behavior, producing test false-alarm rates close to 1.0. Under a validation false-alarm-rate constraint, the GRU forecaster performs best at the evaluated operating point, reaching test F1 of 0.6122, test false-alarm rate of 0.7600, and event recall of 1.0000. The MLP forecaster overfits the threshold and returns to a test false-alarm rate of 1.0000, while the rollout forecaster is more conservative, with test F1 of 0.4390, false-alarm rate of 0.6800, and event recall of 0.6000. These results indicate that transition forecasting contains useful signal, but the operating point remains calibration-sensitive; we therefore do not treat early warning as a main result.

\subsection{Scenario Residual and Error Recovery}
\label{sec:supp_scenario_recovery}

We further analyze whether the learned scores contain information beyond scenario priors. A scenario-only probe obtains 0.7372 Macro-F1 and 0.9089 BAcc. Adding model scores improves the probe to 0.7803 Macro-F1 and 0.9239 BAcc, suggesting that the learned predictions are not reducible to scenario lookup.

We also measure how often the RGB-physiological student corrects RGB-only errors. Among 803 RGB-only errors, 382 are recovered by the RGB-physiological student. Among 646 RGB-only unsafe false negatives, 228 are recovered.

\begin{table}[t]
\centering
\small
\caption{Scenario residual and RGB-error recovery analysis.}
\label{tab:supp_scenario_recovery}
\begin{tabular}{lcc}
\toprule
Analysis & Result 1 & Result 2 \\
\midrule
Scenario-only probe & Macro-F1 0.7372 & BAcc 0.9089 \\
Scenario+score probe & Macro-F1 0.7803 & BAcc 0.9239 \\
RGB-only errors & 803 total & 382 recovered (47.6\%) \\
RGB-only unsafe FNs & 646 total & 228 recovered (35.3\%) \\
\bottomrule
\end{tabular}
\end{table}

\subsection{Matched-Coverage Gate Baselines}
\label{sec:supp_matched_coverage}

To verify that the learned gate is not merely reducing risk by rejecting more samples, we compare policies at the learned gate's coverage of 0.8439. Random abstention and confidence-based thresholding reduce unsafe false negatives less effectively than the learned gate.

\begin{table}[t]
\centering
\small
\caption{Matched-coverage unsafe false-negative comparison. All baselines are evaluated at the learned gate's coverage.}
\label{tab:supp_matched_coverage}
\begin{tabular}{lcc}
\toprule
Policy & Coverage & Unsafe FN \\
\midrule
Always-fast & 1.0000 & 0.1737 \\
Random abstention & 0.8439 & 0.1467 \\
Confidence/entropy/margin threshold & 0.8439 & 0.1027 \\
Learned gate & 0.8439 & \textbf{0.0528} \\
\bottomrule
\end{tabular}
\end{table}

\subsection{Test-Time Physiological Intervention}
\label{sec:supp_physio_intervention}

We evaluate the same strong RGB-physiological checkpoint while replacing the physiological input only at test time. The results show that the model should not be interpreted as relying on exact time-synchronized physiological dynamics.

\begin{table}[t]
\centering
\small
\caption{Test-time physiological intervention using the same strong RGB-physiological checkpoint.}
\label{tab:supp_physio_intervention}
\begin{tabular}{lccc}
\toprule
Test-time physio & Macro-F1 & BAcc & Unsafe FN \\
\midrule
True synchronized & 0.7674 & 0.9221 & 0.1515 \\
Shuffled & 0.7668 & 0.9218 & 0.1521 \\
Shifted4 & 0.7902 & 0.9324 & 0.1309 \\
Shifted16 & 0.7933 & 0.9337 & 0.1283 \\
Shifted64 & 0.7635 & 0.9202 & 0.1552 \\
Zero / missing & \textbf{0.8409} & \textbf{0.9388} & \textbf{0.0837} \\
\bottomrule
\end{tabular}
\end{table}

These results suggest that the current model uses window-level physiological or contextual information, but should not be interpreted as depending on exact temporal physiological synchronization.

\subsection{Abstention Burden}
\label{sec:supp_abstention_burden}

Our main evaluation reports positive abstention and unhandled positives to avoid treating abstention as a free correct prediction. Positive abstention counts true \texttt{high} samples rejected for intervention, while unhandled positives combine unsafe false negatives and positive abstentions. Future evaluations should further separate low-risk false interventions and warning precision when richer intervention logs are available.

% \subsection{Additional Qualitative Results}
% \label{sec:supp_qualitative}

% Additional qualitative examples are provided in Fig.~\ref{fig:supp_modality_cases}--Fig.~\ref{fig:supp_calibration_cases}. We include RGB-only versus RGB-physiological examples, selective confusion visualizations, driver-state rollout cases, and difficult participant-scenario calibration examples. These figures complement the main-paper analysis and illustrate both the benefits and failure modes of the selective monitoring system.

%这个整个从正文移动到附录
\begin{table*}[t]
\centering
\small
\caption{Worst-group and operating-point analysis. The world-model gate improves utility in some settings but remains calibration-sensitive under group shift.}
\label{tab:supp_calibration}
\resizebox{0.9\linewidth}{!}{% 稍微收缩宽度，单列指标表不适合拉得太宽
\begin{tabular}{@{}l cccc@{}}
\toprule
\textbf{Method} & \textbf{Coverage}$_{\uparrow}$ & \textbf{Unsafe FN}$_{\downarrow}$ & \textbf{Unhandled Pos.}$_{\downarrow}$ & \textbf{Utility}$_{\uparrow}$ \\
\midrule
\multicolumn{5}{@{}l}{\textit{Worst-group weighted performance}} \\
Current-evidence gate & -- & \textbf{0.0705} & \textbf{0.1139} & 0.3960 \\
\rowcolor{gray!10}
\textbf{World-model gate ($H=5$, Ours)} & -- & 0.0731 & 0.1246 & \textbf{0.4035} \\
\midrule
\multicolumn{5}{@{}l}{\textit{Matched seed-group pairs (World-model better or equal)}} \\
WM $\geq$ Current & 57/72 & 61/72 & 60/72 & 56/72 \\
\midrule
\multicolumn{5}{@{}l}{\textit{Validation-selected fixed-risk operating point}} \\
Current-evidence gate & 0.9158 & 0.0436 & 0.1225 & -- \\
World-model gate ($H=1$) & 0.9156 & 0.0417 & 0.1241 & -- \\
World-model gate ($H=3$) & 0.9170 & 0.0531 & 0.1317 & -- \\
\rowcolor{gray!10}
\textbf{World-model gate ($H=5$, Ours)} & 0.8933 & \textbf{0.0359} & 0.1428 & -- \\
\midrule
\multicolumn{5}{@{}l}{\textit{Upper bound reference}} \\
\textcolor{gray}{Fixed-risk Oracle} & \textcolor{gray}{\textbf{0.9922}} & \textcolor{gray}{0.0000} & \textcolor{gray}{\textbf{0.0082}} & \textcolor{gray}{--} \\
\bottomrule
\end{tabular}%
}
\end{table*}

\end{document}